\documentclass[conference,10pt]{IEEEtran}
\IEEEoverridecommandlockouts
\usepackage{cite}
\usepackage{amsmath,amssymb,amsfonts,bm}
\usepackage{algorithmic}
\usepackage[ruled,vlined]{algorithm2e}
\usepackage{graphicx}
\usepackage{textcomp}
\usepackage{xcolor}
\def\BibTeX{{\rm B\kern-.05em{\sc i\kern-.025em b}\kern-.08em
    T\kern-.1667em\lower.7ex\hbox{E}\kern-.125emX}}

\begin{document}

\title{
Personalized Federated Learning over non-IID Data for Indoor Localization
}


\author{Peng Wu$^1$,
Tales~Imbiriba$^1$,
Junha Park$^2$,
Sunwoo Kim$^2$,
Pau~Closas$^1$
\\
\IEEEauthorblockA{
$^1$ Electrical and Computer Engineering Department, Northeastern University, Boston, MA \\ \{wu.p,talesim,closas\}@northeastern.edu}
\IEEEauthorblockA{
$^2$  Department of Electronic Engineering, Hanyang University, Seoul, Korea
\\ \{eric0725,remero\}@hanyang.ac.kr
}

\thanks{This work has been partially supported by the NSF under Awards CNS-1815349 and ECCS-1845833.}
}

\maketitle


\begin{abstract}
Localization and tracking of objects using data-driven methods is a popular topic due to the complexity in characterizing the physics of wireless channel propagation models. In these modeling approaches, data needs to be gathered to accurately train models, at the same time that user's privacy is maintained. An appealing scheme to cooperatively achieve these goals is known as Federated Learning (FL). A challenge in FL schemes is the presence of non-independent and identically distributed (non-IID) data, caused by unevenly exploration of different areas. 
In this paper, we consider the use of recent FL schemes to train a set of personalized models that are then optimally fused through Bayesian rules, which makes it appropriate in the context of indoor localization. 
\end{abstract}


\section{Introduction and related works}


There has been a steady increase in 
indoor real-time locating and tracking systems. 
With respect to the broader research domain of Positioning, Navigation and Timing, indoor positioning and tracking offers a number of difficult challenges. One approach to positioning and tracking in such environments is RF fingerprinting 
\cite{dardari2015indoor,wu2019wifi}. The basic idea of fingerprinting is to build a database containing a collection of measured features at designated reference locations within the environment, and to then perform positioning by applying regression or classification techniques to match new measurements to one or more of those in the database. 
Acquiring the database is time consuming and costly, for which crowdsourcing approaches are relevant solutions.  
Traditional crowdsourcing techniques involved users sending their data directly to a centralized server, which in some situation may compromise users’ privacy. Alternatively, Federated Learning (FL), has recently attracted great interest due to its privacy-protecting nature and the efficient use of resources by harnessing the processing power of edge devices\cite{niknam2020federated}. In the context of localization, a number of studies have shown the potential of distributed learning \cite{arias2018crowd,9148111,yin2020fedloc}.



FL is a promising solution that enables many clients to jointly train machine learning models while maintaining local data decentralization. Instead of exchanging data and conducting centralized training, each party sends its model to the server, which updates a joint model and sends the global model back to the parties. Since their original data will not be exposed, FL is an effective way to address privacy issues~\cite{liu2020secure}. 


A key and common data challenge in such distributed approaches is that the data from different clients is often non-independent and identically distributed (non-IID). 
For instance, this is the case when different FL clients collect data from distinct locations. 
According to previous studies \cite{hsu2019measuring, li2019convergence, karimireddy2020scaffold}, non-IID data settings (such as label distribution skewness, feature distribution unbalance, or training data amounts) reduce the effectiveness of learning \cite{li2021federatedLO}.


%


Recent efforts to address FL under non-IID data have been proposed in~\cite{li2020federated,karimireddy2020scaffold,wang2020tackling} but considering a single global objective. However, it is difficult to converge to a global model that provides a good personalized performance on every client when the data is non-IID across different clients. Another branch of studies~\cite{li2021federatedLO,kulkarni2020survey,Fallah2020PersonalizedFL,Ji2021EmergingTI,huang2021personalized,smith2017federated} have addressed the need of personalized local models for each client. Most of the methods\cite{Fallah2020PersonalizedFL,Ji2021EmergingTI}, such as transfer learning or meta-learning, adapt or adjust the global model for individual clients, while some methods\cite{huang2021personalized,smith2017federated} utilize pair-wise collaboration among clients to train personalized models. 





\begin{figure}[!t]
\centerline{\includegraphics[width=80mm]{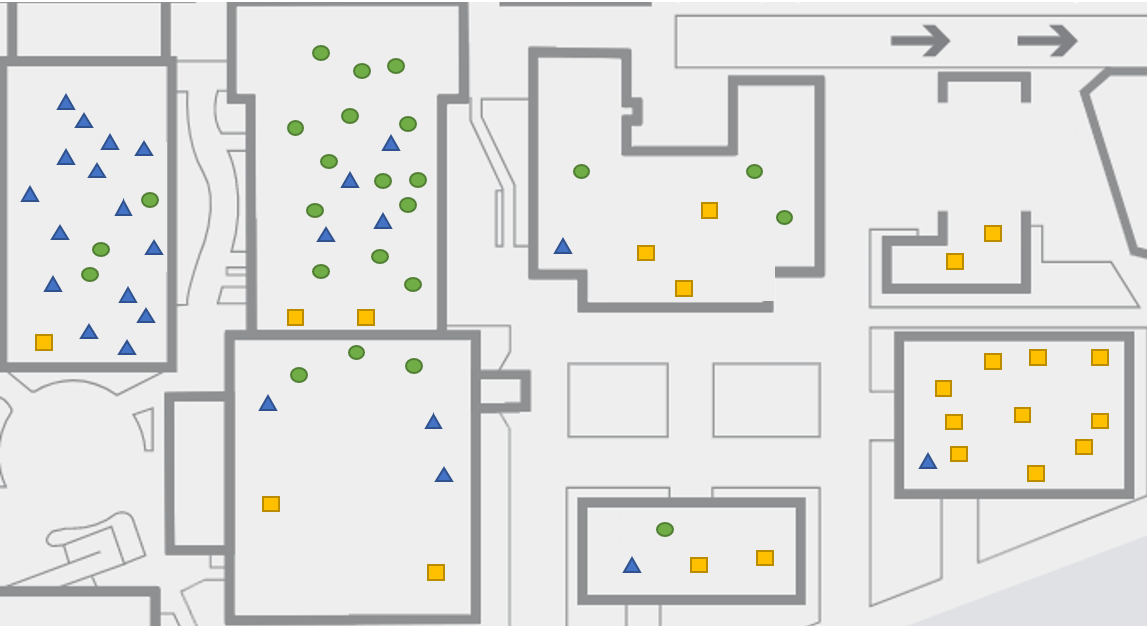}}
\caption{Personalized FL involves clients (or agents) exploring an area in a possibly uncoordinated manner with the goal of gathering features of interest to train local models. 
As a consequence, different clients might observe non-IID data due to exploring different areas more often. In the figure, different markers represent data collection from different agents.}
\label{fig:overview}
\end{figure}

In this paper we envisage the use of personalized FL models for localization purposes. In particular, we consider crowdsourced received signal strength (RSS) of WiFi data to build fingerprinting (data-driven) models, an approach that does not compromise training user's privacy. 
This problem connects with several scenarios where clients might be moving in large, and possibly intricate, areas such as market halls, hospitals, university campus, or even entire cities. 
In such scenarios the non-IID nature of the collected data is evident, which is illustrated in Fig.~\ref{fig:overview} with different clients focusing on different areas.
The inherent complexity of local buildings and signal propagation makes personalized models appealing solutions that can obtain region-specific models and at the same time exploit information obtained from other areas. For this reason, we leverage Personalized FL strategies formulating the localization problem as a classification problem. We specially focus on the recent algorithm FedAMP \cite{huang2021personalized} since it shows superior classification performance in comparison to related algorithms. Although FedAMP obtained good performance on different classification problems the direct application to localization problems requires some adjustments. This is so, due to the data association problem between data acquired in an unknown location and all possible personalized models. To address this issue, we propose a Bayesian fusion strategy that combines the result of multiple personalized models into a unique global model during the test phase. 

\section{Federated Learning schemes}

The idea behind the FL is to approximate a global model as an integration of local models. Mathematically it can be formulated as the minimization of a loss function:

\begin{equation}\label{eq:FL}
\min_\mathbf{W} \mathcal{L}(\mathbf{W}) \text{     where    } \mathcal{L}(\mathbf{W}) = \sum_{i=1}^{M} \mathcal{F}_i(\mathbf{w}_i)
\end{equation}
where $\mathbf{W} = [\mathbf{w}_1, \ldots, \mathbf{w}_M] \in \mathbb{R}^{d\times M}$ contains the model parameters,  $\mathcal{L}(\mathbf{W})$ is the global cost functional, while $\mathcal{F}_i: \mathbb{R}^d\rightarrow \mathbb{R}, \mathbf{w}_i \mapsto \mathcal{F}_i(\mathbf{w}_i)$ 
are local cost functions trained using disjoint datasets $\mathcal{D}_1, \ldots, \mathcal{D}_M$. 

Among the different strategies to solve~\eqref{eq:FL} we highlight the conventional FL approach: FedAvg, which considers a single global objective, as well as the latest personalized approach FedAMP, which considers several personalized local models. 

\subsection{FedAvg}
Federated averaging (FedAvg) is a \textit{de facto} approach for FL in which local models are trained locally. The clients then upload their models to a cloud server, which is in charge of fusing the model parameters (e.g. weights in a neural network) to compute a global model. 
In order to account for each local model quality, \eqref{eq:FL} can be modified as:
\begin{equation}\label{eq:Fedavg}
\mathcal{L}(\mathbf{W}) = \sum_{i=1}^{M} \frac{N_i}{N} \mathcal{L}_i(\mathbf{w}_i)
\end{equation}
where $\mathcal{L}_{i}(\mathbf{w}_i)=\frac{1}{N_{i}} \sum_{n \in \mathcal{P}_{i}} f_{n}(\mathbf{w}_i)$, $f_n(\mathbf{w}_i)$ is the loss of the prediction on sample $n$ made with model parameters $\mathbf{w}_i$. $\mathcal{P}_{i}$ is the data partition for client $i$, $N$ is the total number of data points and $N_i$ is the number samples used by the $i$-th client.
FedAvg is an efficient FL scheme, but may lead to poor accuracy results in certain situations such as non-IID cases, as shown in previous studies~\cite{hsu2019measuring,li2019convergence}. 



\subsection{FedAMP}

The federated attentive message passing (FedAMP) algorithm was devised as a privacy-preserving personalized cross-silo federated learning strategy for non-IID data \cite{huang2021personalized}. It aims at solving an optimization problem of the form:
\begin{equation}\label{eq:personalizedFED}
\min_\mathbf{W} \mathcal{L}(\mathbf{W}) +  \lambda \mathcal{R} (\mathbf{W})
\end{equation}
where $\lambda\in\mathbb{R}_+$ and $\mathcal{R} (\mathbf{W}) = \sum_{i\neq j} h (\|\mathbf{w}_i - \mathbf{w}_j\|^2)$ is a regularization function composed of continuously differentiable attention-inducing functions $h: [0,\infty] \rightarrow \mathbb{R}$ aimed at constraining the $M$ models to have \textit{similar} parameters. The problem in~\eqref{eq:personalizedFED} was reformulated by exploiting incremental gradient proximal methods to sequentially estimate the personalized model parameters $\mathbf{W}$, by solving 
\begin{equation}
    \mathbf{W}^k = \arg\min_{\mathbf{W}} \mathcal{L}(\mathbf{W}) + \tilde{\lambda} \|\mathbf{W} - \mathbf{U}^k\|^2_F
\end{equation}
at every iteration $k=1,\ldots, K$, where $\mathbf{U}^k = [\mathbf{u}_1^k, \ldots, \mathbf{u}_m^k]$ is used as the prox-center~\cite{huang2021personalized}, and can be updated as
\begin{equation}\label{eq:U_update}
    \mathbf{U}^k = \mathbf{W}^{k-1} - \alpha_k \nabla \mathcal{R} (\mathbf{W}^{k-1}).
\end{equation}
In~\eqref{eq:U_update} $\alpha_k\in\mathbb{R}_+$ is the step size of the gradient descent method and $\tilde{\lambda} = \lambda/2\alpha_k$, where $\lambda\in\mathbb{R}_+$ is a parameter controlling the cross-silo regularization. It has been shown~\cite{huang2021personalized} that
$\mathbf{u}_i^k\in\mathbb{R}^d$ computed in~\eqref{eq:U_update} can be written as a convex combination of the model parameters as:
\begin{equation}
    \mathbf{u}_i^k = \xi_{i,1}\mathbf{w}_1^{k-1} + \cdots + \xi_{i,M}\mathbf{w}_M^{k-1}
\end{equation}
where $\xi_{i,i^\prime}\in\mathbb{R}_+$, with $\sum_{i^\prime} \xi_{i,i^\prime} = 1$, can be interpreted as similarity coefficients due to the properties of the attention function $h$~\cite{huang2021personalized}.

Important features of the methodology proposed by Huang et al.~\cite{huang2021personalized}, briefly discussed above, include privacy preserving, since data used to learn local models are never shared with other nodes, and collaborative learning using an attentive message passing mechanism which ultimately induces local models with high similarity coefficient to collaborate. 
These features are extremely important for distributed indoor localization problems since data can be scarce and privacy is crucial, specially in scenarios where local models from different areas (or buildings) might be exchanging information. This is even more so, if considering large scale scenarios where models might be learned over multiple neighbourhoods or even over entire map. 

Although this strategy proved relevant for different classification problems~\cite{huang2021personalized} its direct application to indoor positioning is not so direct, presenting challenges regarding data distribution and model prediction. Regarding data collection and distribution we will assume that data used to train a particular model can come from any location but will be more concentrated at the region associated with a given local model. Regarding the prediction step, since data arriving comes from an unknown location we propose to fuse multiple local models to provide a final prediction. This will be addressed in more detail in the next section.



\section{Fusion of Federated Learning models}~\label{sec:Fusion}


The FedAMP approach discussed earlier produces a set of $M$ models that are well fitted to the data they were trained with. That is, if data was primarily from certain areas in the map, the predictive accuracy of those models is expected to be larger in those areas. Conversely, worse in those areas where data quantity is smaller than other personalized models. As a consequence, a user utilizing the models has no a priori knowledge which is the correct model to use for localization purposes (that is precisely the objective of the models). Therefore, we propose a fusion scheme that blends the $M$ available models to produce a localization estimate. Here we are considering that the localization task involves determining which area (e.g. room, hall, etc.) the user is in, given the measured RSS data. 

In the FedAMP scheme, the $i \in \{1,\dots,M\}$ classifier (i.e. each of the personalized models) provides a categorical posterior distribution for $c$, the class, which can take any of the $j \in \{1,\dots,L\}$ labels (i.e. each room or area) based on the current data $\mathbf{y}$ and the corresponding model $\mathcal{M}_i$, that is
\begin{equation}
    p\left(c|\mathcal{M}_i, \mathbf{y}\right) = \prod_{j=1}^L p_{j|i}^{[c=j]},
\end{equation}
where $p_{j|i}$ denotes the probability for the $j$-th label given the $i$-th classifier, and $[c=j]$ is an indicator function that returns $1$ if $c=j$ and $0$ otherwise. 
The a priori class probability $p\left( c \right)$ is categorical and is defined by the probabilities $p_{j|0}$, which in the equiprobable case result in $p_{j|0}=1/L$, $\forall j$. 
The optimal fusion rule is provided by the joint posterior distribution, which can be shown \cite{pastor2020bayesian} to be proportional to
\begin{align}
    \notag p\left( c | \mathcal{M}_{1:M}, \mathbf{y} \right) & \propto 
    \frac{\prod_{i=1}^M p\left( c | \mathcal{M}_{i}, \mathbf{y} \right)}{ p(c) }
    =\frac{\prod_{i=1}^M \prod_{j=1}^L (p_{j|i})^{[c=j]}} { p(c) } \\
    &= \prod_{j=1}^L {\left( \frac{p_{j|1} \cdot p_{j|2} \cdots p_{j|M} }{ p_{j|0}^{M-1} } \right)}^{[c=j]} \;,
\end{align}
\noindent where we have used that the different models are conditionally independent given $c$. The resulting joint distribution is categorical, from which the maximum a posteriori (MAP) can be readily obtained to predict the class $c$. 

\section{Positioning Experiments}
\begin{figure}[htbp]
\centerline{\includegraphics[width=0.95\columnwidth]{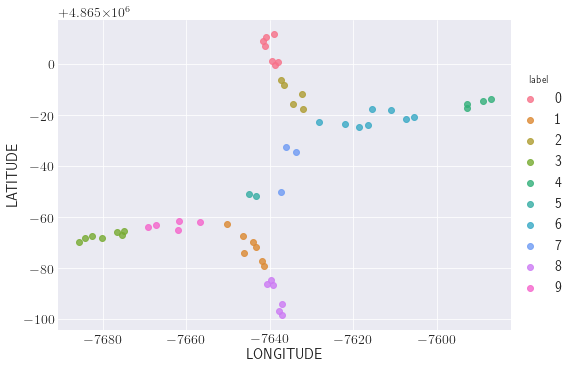}}
\caption{Latitude/Longitude map of the data location in the floor under test. Different color represent clusters of nearby data, treated as a single label in this study.}
\label{room_cluster}
\end{figure}

For our experiments we use the public available RSS data from UJIIndoorLoc database \cite{7275492} is used to evaluate our proposed federated learning-based localization
system. The UJIIndoorLoc database includes 19937 RSS samples distributed over 390 meters in length
to 270 meters in width area over four floors for training and testing. In our experiments we focus on floor 1 located at building 1 to reduce the computational cost. 
Figure~\ref{room_cluster} shows the latitude and longitude of the rooms in the selected floor. 
Because some rooms had very few samples, we grouped neighbouring rooms to form larger areas depicted in Figure~\ref{room_cluster} by color and labeled from 0 to 9. Our experiments will focus in classifying RSS data coming from these 10 areas, but first we will present our non-IID data spliting strategy. 

\subsection{Settings of non-IID Datasets}

Among different non-IID possibilities, we focus here on having an unbalanced distribution of the labels.
For example, some clients collect more specialized points in several specific areas after navigating the scene more often. 
To simulate label distribution skewness across clients, we use the Dirichlet distribution. This method has been used in many recent FL studies \cite{li2021federatedLO}
Specifically, for a given client $i$, we define the probability of sampling data from the $j \in \{1,\dots,L\}$ label as the vector $(p_{i,1},\dots,p_{i,L}) \sim \operatorname{Dir}(\bm{\beta})$, where $\operatorname{Dir}(\cdot)$ denotes the Dirichlet distribution and $\bm{\beta} = (\beta_{i,1}, \dots, \beta_{i,L})$ is the concentration vector parameter ($\beta_{i,j}>0$, $\forall j$). 
The advantage of this approach is that the imbalance level can be flexibly changed by adjusting the concentration parameter $\beta_{i,j}$. If it is set to a smaller value, the partition is more unbalanced.

\begin{figure}[htbp]
\centerline{\includegraphics[width=0.95\columnwidth]{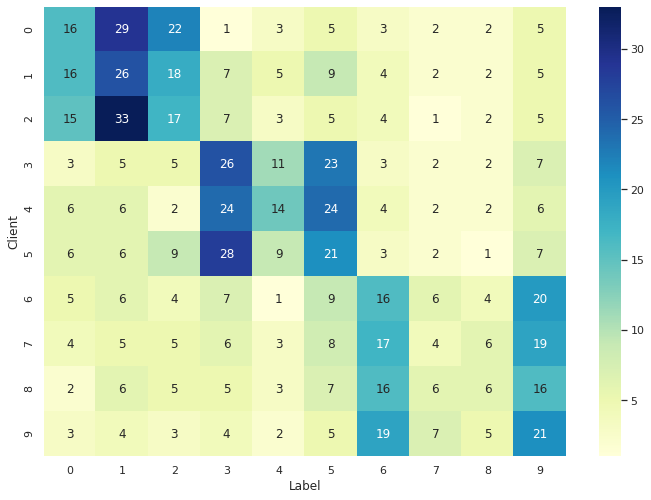}}
\caption{An example of distribution label imbalance partition on the dataset. The value in each rectangle represents the number of data samples of a given label belonging to a given client. In this example, 3 groups of clients can be observed, each having similarly distributed data but non-IID across groups.}
\label{fig:non_iid}
\end{figure}

However, unlike other works, we are not only interested in generating some unbalanced data proportion, but also want that some clients have a similar label distribution, similarly to \cite{huang2021personalized}. 
Fig. \ref{fig:non_iid} shows an example non-IID distribution with a total of 10 labels and 10 clients gathering data in the area. A total of 3 groups are considered, each group containing 3 dominating labels.  
In this setup, we assign the dominating classes in first group by sampling data for clients $0-2$ using large $\beta$ values in those labels $\bm{\beta} = (80,80,80,20,\dots,20)$.  
Similar strategy is used when assigning the dominating classes in other groups of clients ($3-5$ and $6-9$ in the example). 
Using the aforementioned Dirichlet sampling approach provides more randomness to the study, as opposite to setting a fix number of data points per client. 

\subsection{Results}
In this section we compare five different strategies that differ in their training and testing settings. Specifically, we consider Global Model (GM) and Local Models (LMs) where the first is trained using data from all areas whereas for LMs each individual model is trained using only non-IID data associated to its specific area(s). Those models are neural networks (NN) with the same MLP structure, having two hidden layers with $[256,16]$ neurons and ReLu activations.  
Furthermore, as discussed in Section~\ref{sec:Fusion} we perform fusion of the LMs class predictions to generate a unique prediction. We call this LM with Fusion (LM-F). As a comparison we also present results when LMs are directly used for classification of RSS values. In this case the accuracy is computed averaging the accuracy obtained for each local model. The same applies for the FL models FedAvg, FedAMP and the fused version of the latter: FedAMP-F. Notice that FedAvg does not have fused version because it is already based on building a unique global model. 
For all experiments we performed  
50 Monte Carlo runs (randomly initializing the NNs and performing data shuffling) to make the results more stable and reliable.

\begin{figure}[htbp]
\centerline{\includegraphics[width=0.95\columnwidth]{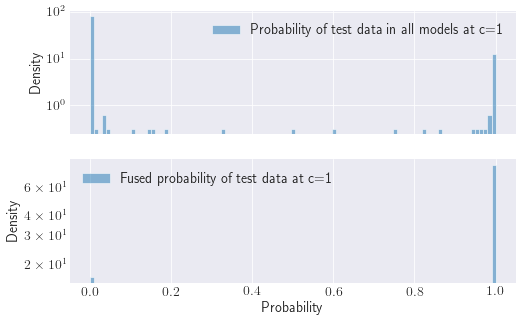}}
\caption{Distribution of classification probabilities (top) before and (bottom) after fusion of the different local models.}
\label{fig:prbability_density0}
\end{figure}

Figure~\ref{fig:prbability_density0}, shows how the probability distribution changes before (FedAMP) and after fusion (FedAMP-F) is performed. Specifically, when class $c=1$ is to be classified, the top panel shows the probability distribution of $p_{1|i}$, $i=1,\ldots, M$. 
We can observe a large performance variability depending on the used model. 
The bottom panel shows the distribution of $p\left( c=1 | \mathcal{M}_{1:M}, \mathbf{y} \right)$ after fusion, which shows clear improvements in terms of classifying correctly the label.

\begin{figure}[htbp]
\centerline{\includegraphics[width=0.95\columnwidth]{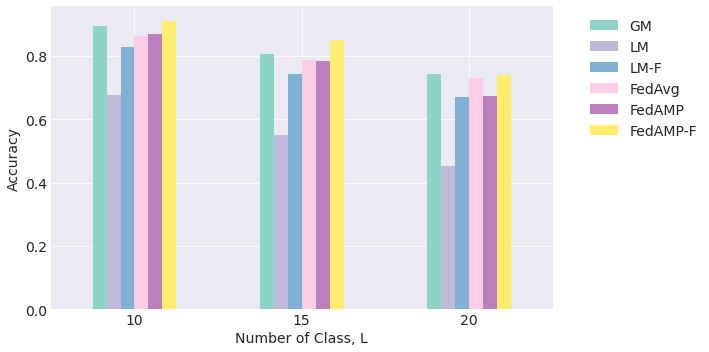}}
\caption{Accuracy of models as a function of $L$, the number of classes.}
\label{fig:accuracy_class}
\end{figure}

\begin{figure}[htbp]
\centerline{\includegraphics[width=0.95\columnwidth]{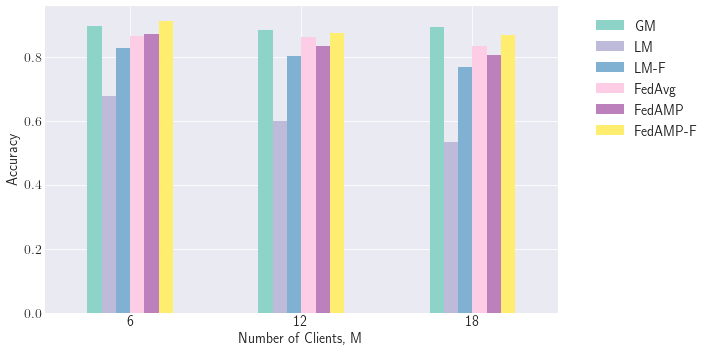}}
\caption{Accuracy of models as a function of $M$, the number of clients.}
\label{fig:accuracy_client}
\end{figure}

We present performance results in Figures~\ref{fig:accuracy_class} and~\ref{fig:accuracy_client} where GMs are used merely for benchmarks purposes. 
Fig. \ref{fig:accuracy_class} shows the accuracy of all methods under the non-IID data setting and different number of classes $L$ while keeping the number of clients fixed $M=6$. Analyzing the figure we observe that, for all models, the performance decreases as the number of classes increases. This is expected since increasing $L$ reduces the number of training samples available for each region, $N_i$. When comparing the different non-global strategies we highlight the poor accuracy of LM models. This is so, due to the data association problem and the lack of collaboration among clients. However, large improvements are achieved when introducing client collaboration (i.e., FL methods) or/and fusion. For instance, LM-F produces and average accuracy improvement in comparasison with LM of approximately  18.7\% while FedAvg and FedAMP leads to average improvements of 23.4\% and 21.7\%, respectively. The best local model results were obtained when considering FL strategies in conjunction with the Fusion strategy. This can be seen for all number of classes where FedAMP-F presents the best non-global result. In fact, for $L=10$ and $L=15$ FedAMP-F performs even better than the GM. This is due, however, to the fact that the same NN structure is used for GM and LMs giving local models more flexibility.    
%
%
%
In Fig.~\ref{fig:accuracy_client} we also examine the performance variation across different number of clients, $M$, while keeping the number of classes fixed, $L=10$. As the number of clients increase, the accuracy of LMs and LM-Fs decrease. This behavior is also expected since increasing the number of clients also reduces the amount of data available for training each client. While FedAvg achieves relatively stable results, the accuracy of FedAMP is decreasing as $M$ increases. However, fusing the local FedAMP models (FedAMP-F) leads to the best results. 

\subsection{Results for different hyper-parameters settings}
\begin{figure}[htbp]
\centerline{\includegraphics[width=0.95\columnwidth]{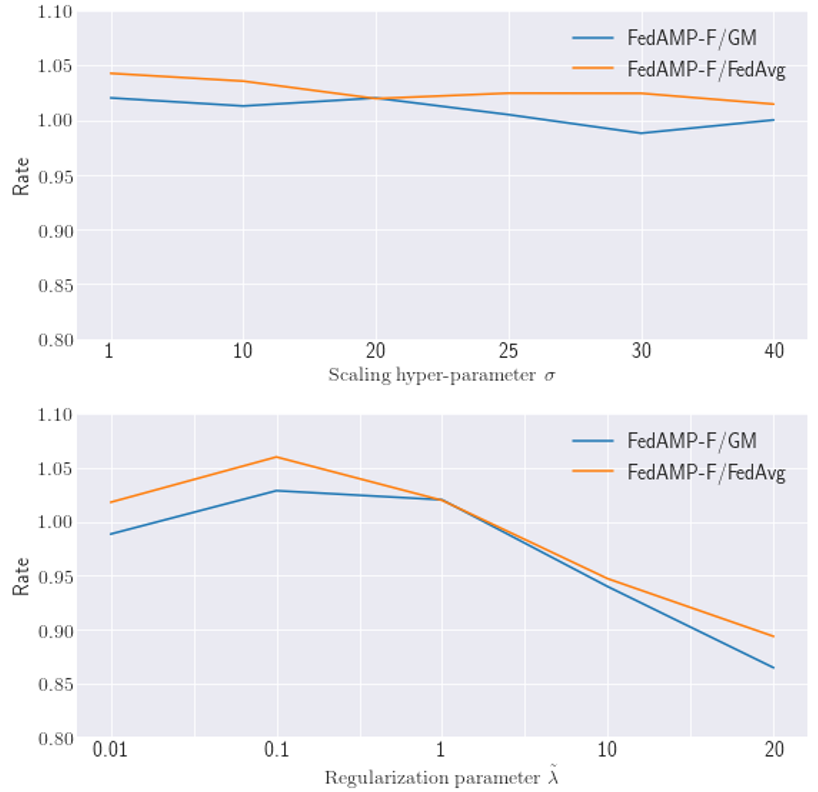}}
\caption{Sensitivity analysis of FedAMP-F to (top) $\sigma$ and (bottom) $\tilde{\lambda}$.} 
\label{fig:hyperparameters}
\end{figure}

In this section we examine the performance of the FedAMP and FedAMP-F as function of their hyperparameters: $\tilde{\lambda}$ and $\sigma$.
Fig.~\ref{fig:hyperparameters} shows relative performance curves where ``Rate'' is the ratio between the accuracy of FedAMP-F, GM and FedAvg as described in the corresponding legends. 
Observing the top panel we can see that FedAMP-F seems quite insensitive to variations of $\sigma$.
The bottom panel, however, shows great impact on the FedAMP-F performance when varying $\tilde{\lambda}$. The best results where obtained in the range $[0.01, 1]$.
The results in Figures~\ref{fig:accuracy_class} and~\ref{fig:accuracy_client} were obtained with $\sigma=20$, $\tilde{\lambda}=1$.

\section*{Conclusions}


In this paper, we tackle the challenging problem of non-IID data in indoor positioning by proposing a Bayesian fusion rule to merge personalized 
Federated Learning models. This solution enables the applicability of such training schemes in situations where we do not have prior knowledge regarding which model to use. 
We show how the fused model can remarkably improve the performance with respect to personalized models. 






\bibliographystyle{IEEEtran}
\bibliography{IEEEabrv,biblio}



\end{document}